\documentclass[11pt,a4paper]{article}
\usepackage[hyperref]{acl2019}
\usepackage{times}
\usepackage{latexsym}
\usepackage[encapsulated]{CJK}
\usepackage{graphicx} 
\usepackage{float} 
\usepackage{subfigure} 
\usepackage{color}
\usepackage{booktabs}
\usepackage{url}
\usepackage{amsmath}
\usepackage{multirow}
\aclfinalcopy 

\title{Lattice-Based Transformer Encoder for Neural Machine Translation}

\author{Fengshun Xiao$^{1,2}$, Jiangtong Li$^{2,3}$, Hai Zhao$^{1,2,}$\thanks{$\ $ Corresponding author. This paper was partially supported by National Key Research and Development Program of China (No. 2017YFB0304100) and key projects of National Natural Science Foundation of China (No. U1836222 and No. 61733011). Rui Wang was partially supported by JSPS grant-in-aid for early-career scientists (19K20354): ``Unsupervised Neural Machine Translation in Universal Scenarios" and NICT tenure-track researcher startup fund ``Toward Intelligent Machine Translation".}, Rui Wang$^{4}$, Kehai Chen$^{4}$ \\
$^{1}$Department of Computer Science and Engineering, Shanghai Jiao Tong University \\
$^{2}$Key Laboratory of Shanghai Education Commission for Intelligent Interaction \\ and Cognitive Engineering, Shanghai Jiao Tong University, Shanghai, China\\
$^{3}$College of Zhiyuan, Shanghai Jiao Tong University, China\\
$^{4}$National Institute of Information and Communications Technology (NICT) \\
{\tt \{felixxiao, keep\_moving-lee\}@sjtu.edu.cn, } \\
{\tt zhaohai@cs.sjtu.edu.cn, \{wangrui, khchen\}@nict.go.jp}
}

\date{}

\begin{document}
\begin{CJK*}{UTF8}{gbsn}
\maketitle
\begin{abstract}
Neural machine translation (NMT) takes deterministic sequences for source representations. However, either word-level or subword-level segmentations have multiple choices to split a source sequence with different word segmentors or different subword vocabulary sizes. We hypothesize that the diversity in segmentations may affect the NMT performance. To integrate different segmentations with the state-of-the-art NMT model, Transformer, we propose lattice-based encoders to explore effective word or subword representation in an automatic way during training. We propose two methods: 1) lattice positional encoding and 2) lattice-aware self-attention. These two methods can be used together and show complementary to each other to further improve translation performance. Experiment results show superiorities of lattice-based encoders in word-level and subword-level representations over conventional Transformer encoder.
\end{abstract}

\section{Introduction}

\begin{figure}[] 
\centering 
\includegraphics[width=0.48\textwidth]{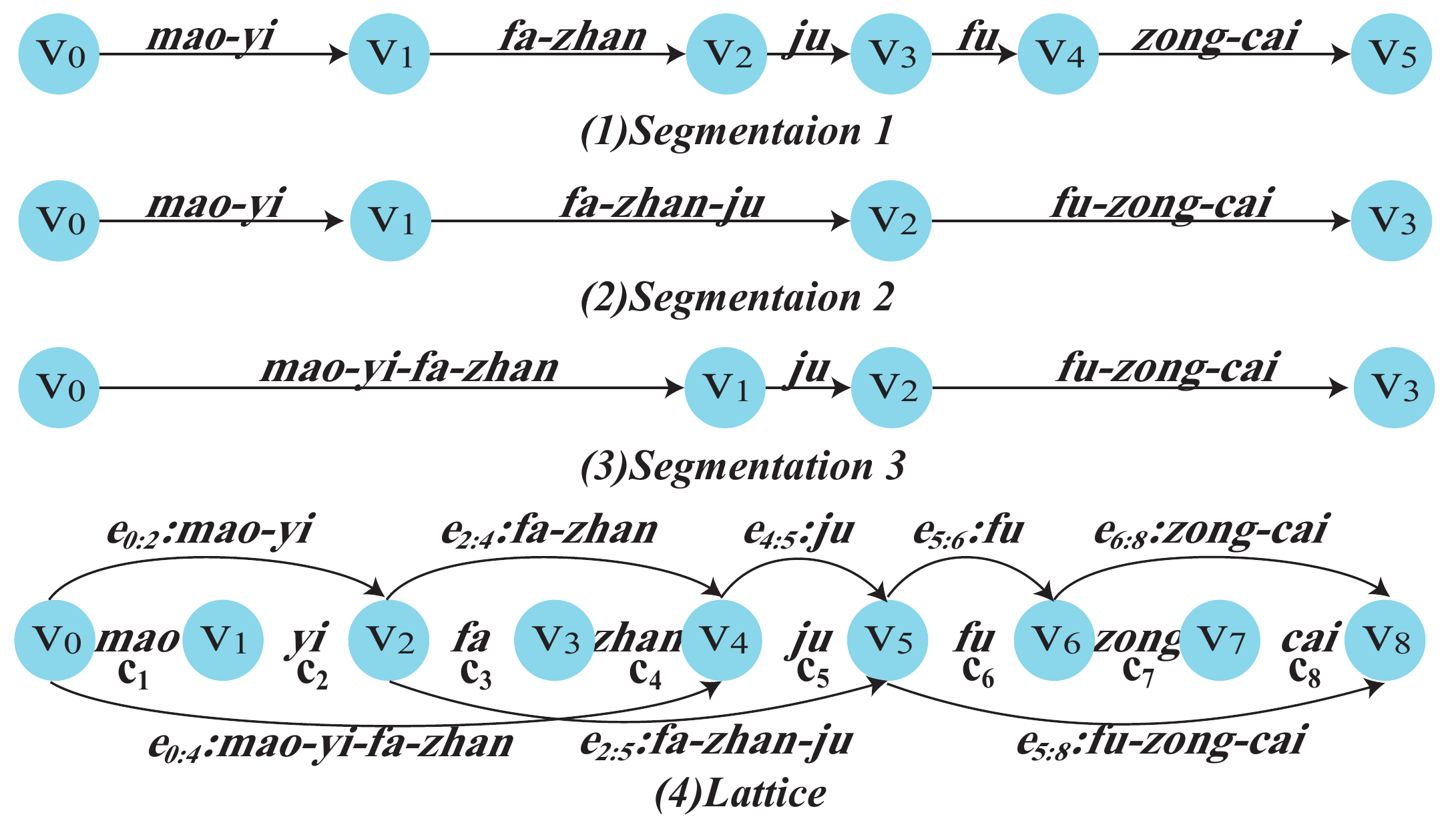}
 \caption{Incorporating three different segmentation for a lattice graph. The original sentence is ``\emph{mao-yi-fa-zhan-ju-fu-zong-cai}". In Chinese it is ``\emph{贸易发展局副总裁}". In English it means ``\emph{The vice president of Trade Development Council}"} 
\label{lattice} 
\end{figure}

Neural machine translation (NMT) has achieved great progress with the evolvement of model structures under an encoder-decoder framework~\cite{sutskever2014sequence,bahdanau2014neural}. Recently, the self-attention based Transformer model has achieved state-of-the-art performance on multiple language pairs \cite{vaswani2017attention,marie-EtAl:2018:WMT}. Both representations of source and target sentences in NMT can be factorized in character \cite{costa2016character}, word \cite{sutskever2014sequence}, or subword \cite{sennrich2015neural} level. However, only using 1-best segmentation as inputs limits NMT encoders to express source sequences sufficiently and reliably. Many East Asian languages, including Chinese are written without explicit word boundary, so that their sentences need to be segmented into words firstly \cite{zhao2019chinese,cai2017fast,cai2016neural,zhao2013empirical,zhao2011integrating}. By different segmentors, each sentence can be segmented into multiple forms as shown in Figure \ref{lattice}. Even for those alphabetical languages with clear word boundary like English, there is still an issue about selecting a proper subword vocabulary size, which determines the segmentation granularities for word representation. 

In order to handle this problem, \citet{morishita2018improving} used hierarchical subword features to represent sequence with different subword granularities. \citet{su2017lattice} proposed the first word-lattice based recurrent neural network (RNN) encoders which extended Gated Recurrent Units (GRUs) \cite{cho2014learning} to take in multiple sequence segmentation representations. \citet{sperber2017neural} incorporated posterior scores to Tree-LSTM for building a lattice encoder in speech translation. All these existing methods serve for RNN-based NMT model, where lattices can be formulized as directed graphs and the inherent directed structure of RNN facilitates the construction of lattice. Meanwhile, the self-attention mechanism is good at learning the dependency between characters in parallel, which can partially compare and learn information from multiple segmentations \citep{D18-1461}. Therefore, it is challenging to directly apply the lattice structure to Transformer. 

In this work, we explore an efficient way of integrating lattice into Transformer. Our method can not only process multiple sequences segmented in different ways to improve translation quality, but also maintain the characteristics of parallel computation in the Transformer.



\section{Background}

\subsection{Transformer}

Transformer stacks self-attention and point-wise, fully connected layers for both encoders and decoders. Decoder layers also have another sub-layer which performs attention over the output of the encoder. Residual connections around each layer are employed followed by layer normalization \cite{ba2016layer}.

To make use of the order of the sequence, \citet{vaswani2017attention} proposed Positional Encodings to indicate the absolute or relative position of tokens in input sequence which are calculated as:
 
$\quad \quad \quad \quad p_{(j,2i)}=sin(j/10000^{2i/d})$ 

$\quad \quad \quad p_{(j,2i+1)}=cos(j/10000^{2i/d}),$ \\
where $j$ is the position, $i$ is the dimension and $d$ is the model dimension. Then positional encodings $p_{1:M}=\{p_1,...,p_M\}$ are added to the embedding of each token $t_{1:M}=\{t_1,...,t_M\}$ and are propagated to higher layers via residual connections.

\subsection{Self-Attention}

Transformer employs $H$ attention heads to perform self-attention over a sequence individually and finally applies concatenation and linear transformation to the results from each head, which is called multi-head attention \cite{vaswani2017attention}. Every single head attention in multi-head attention is calculated in a scaled dot product form:

\begin{equation}\label{1}
    u_{ij} = \frac{(t_iW^Q)(t_jW^K)^T}{\sqrt{d}},
\end{equation}%
where $d$ is the model dimension, $t_{1:M}$ is the input sequence and $u_{ij}$ are normalized by a softmax function:

\begin{equation}\label{2}
    \alpha_{ij} = \frac{\exp(u_{ij})}{\sum^M_{k=1} \exp(u_{ik})},
\end{equation}%
and $\alpha_{ij}$ are used to calculate the final output hidden representations:

\begin{equation}\label{3}
    o_i = \sum^M_{j=1}\alpha_{ij}(t_jW^V),
\end{equation}%
where $o_{1:M}$ is outputs and $W^Q$,$W^K$, and $W^V$ are learnable projections matrices for query, key, and value in a single head, respectively.



\begin{table}[t] \small
    \centering
    \begin{tabular}{|c|c|c|} \hline
     & \textbf{Conditions} & \textbf{Explanation}\\ \hline \hline
     \textbf{lad} & $ i<j=p<q$ & $e_{i:j}$ is left adjacent  to $e_{p:q}$. \\ \hline
     \textbf{rad} & $p < q = i < j$ & $e_{i:j}$ is right adjacent to $e_{p:q}$. \\
     \hline
     \textbf{inc} & $i \le p < q \le j$ &  $e_{i:j}$ includes  $e_{p:q}$. \\ \hline
     \textbf{ind} & $ p \le i < j \le q$ &  $e_{i:j}$is included  in $e_{p:q}$. \\ \hline
     \multirow{2}{*}{\textbf{its}} & $i < p < j < q$ or &  \multirow{2}{*}{$e_{i:j}$ is intersected  with $e_{p:q}$.} \\ 
     &  $p < i < q < j$ & \\ \hline
     \textbf{pre} &  $i < j < p < q$ & $e_{i:j}$ is preceding edge to $e_{p:q}$. \\ \hline
     \textbf{suc} &  $p < q < i < j$&  $e_{i:j}$ is succeeding edge to $e_{p:q}$. \\ \hline
    \end{tabular}
    \caption{Relations possibly satisfied by any two different edges $e_{i:j}$ and $e_{p:q}$ in the lattice. Note that two equal signs cannot stand at the same time in condition inequality for \textbf{inc} and \textbf{ind}.}
    \label{relation}. 
\end{table}

\section{Models}

\subsection{Lattices}


Lattices can represent multiple segmentation sequences in a directed graph, as they merge the same subsequence of all candidate subsequences using a compact way. 

As shown in Figure \ref{lattice}, we follow \citet{su2017lattice} to apply different segmentator to segment an element\footnote{Character for word lattice and minimum subword unit in our predefined subword segmentations for subword lattice.} sequence $c_{1:N}=\{c_1,c_2,...,c_N\}$  into different word or subword sequences to construct a lattice $G=  \langle V ,E  \rangle$, a directed, connected, and acyclic graph, where $V$ is node set and $E$ is edge set, node $v_i \in V$ denotes the gap between $c_i$ and $c_{i+1}$, edge $e_{i:j} \in E$ departing from $v_i$ and arrives at $v_j$ ($i < j$) indicates a possible word or subword unit covering subsequence $c_{i+1:j}$.

All the edges in the lattice $G$ are the actual input tokens for NMT. For two different edges $e_{i:j}$ and $e_{p:q}$, all possible relations can be enumerated as in Table \ref{relation}.



\subsection{Lattice-Based Encoders}

\begin{figure}[t] 
\centering 
\includegraphics[width=0.48\textwidth]{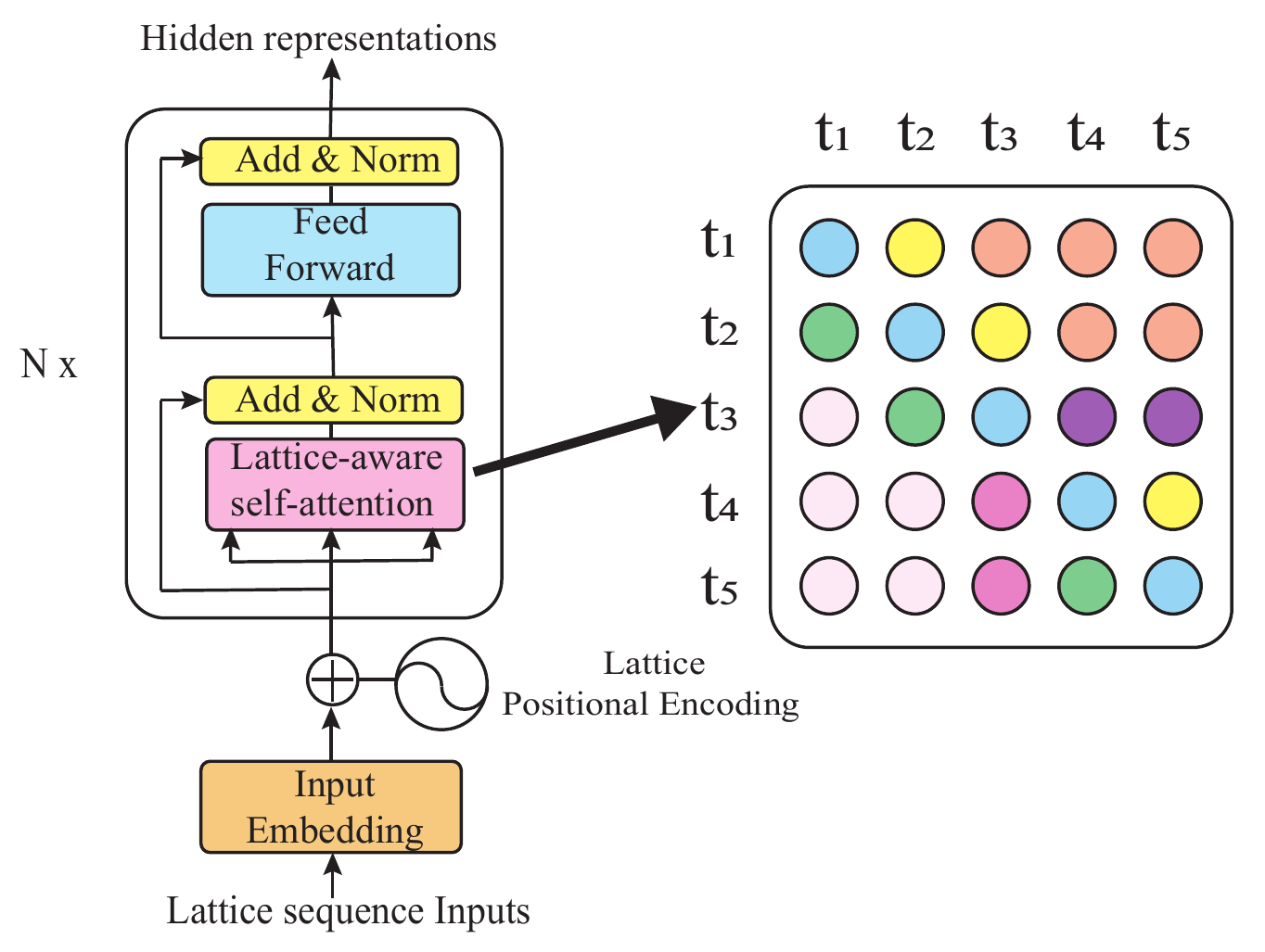}
\caption{The architecture of lattice-based Transformer encoder. Lattice positional encoding is added to the embeddings of lattice sequence inputs. Different colors in lattice-aware self-attention indicate different relation embeddings.} 
\label{arch} 
\end{figure}

We place all edges $E$ in the lattice graph into an input sequence $t_{1:M}=\{t_1,t_2,...,t_M\}$ for Transformer; then we modify the positional encoding to indicate the positional information of input tokens, namely all edges in the lattice graph. In addition, we propose a lattice-aware self-attention to directly represent position relationship among tokens. The overall architecture is shown in Figure \ref{arch}.

\paragraph{Lattice Positional Encoding (LPE)}

\begin{figure}[] 
\centering 
\includegraphics[width=0.48\textwidth]{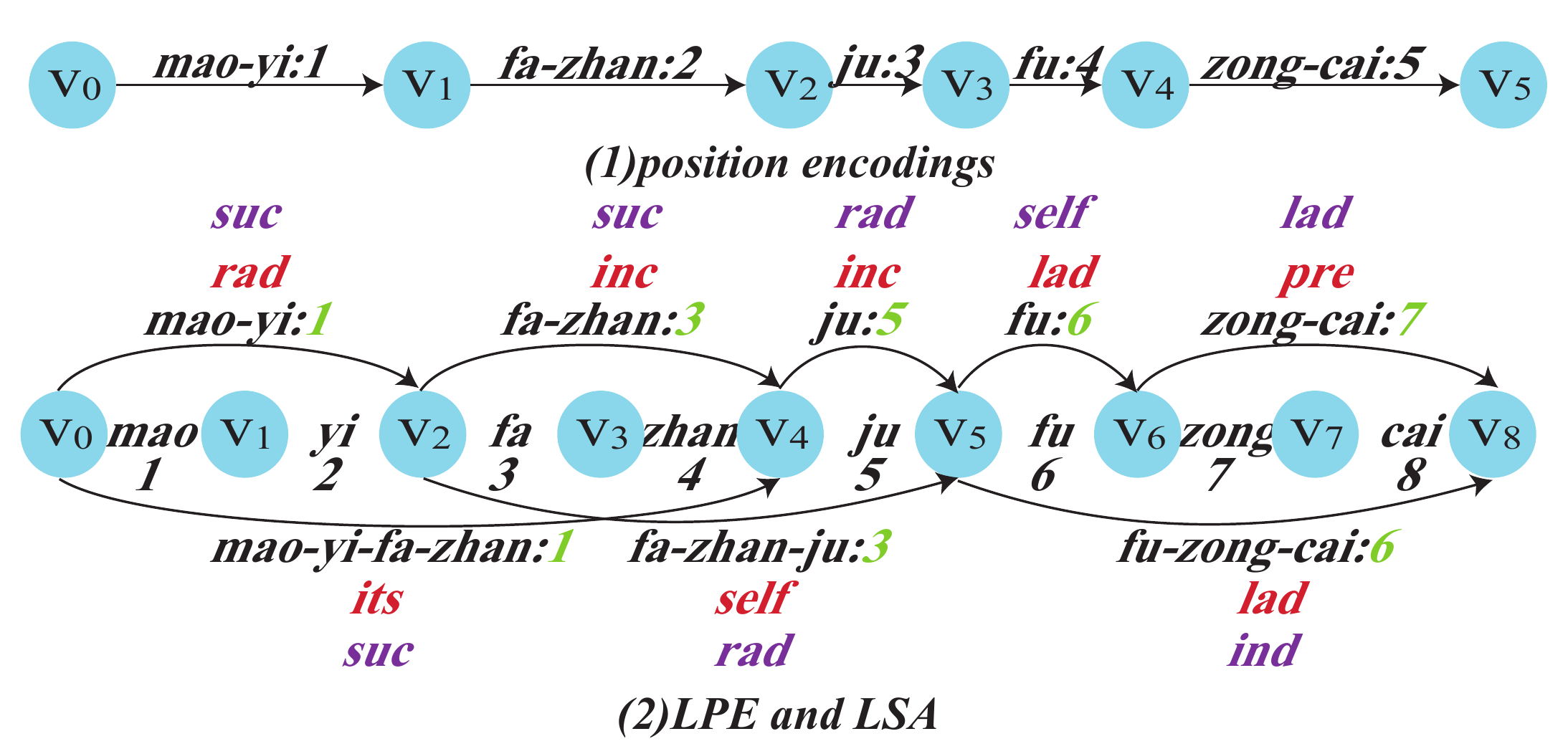}
\caption{Lattice positional encoding $p_{i+1}$ (in green) for edge $e_{i:j}$ in the lattice graph and the relation embeddings $r$ in lattice-aware self-attention based on the timestep of token \emph{fa-zhan-ju} (in red) and \emph{fu} (in purple).} 
\label{lpe} 
\end{figure}

Original positional encoding indicates the order of the sequence in an ascending form $\{p_1,p_2,...,p_M\}$. We hypothesize that increasing positional encodings can indicate the order of sequential sentence. As shown in Figure \ref{lpe}, we scan a source sequence by element $c_{1:N}=\{c_1,c_2,...,c_N\}$ (for example, $c_i$ is character in Figure \ref{lpe}) and record their position $p_{1:N}=\{p_1,p_2,...,p_N\}$. Then we use the positional encoding of the first element in lattice edge to represent current token's position, which can ensure that every edge in each path departing from $v_0$ and arriving at $v_N$ in lattice will have an increasing positional encoding order. The property mentioned above is easy to prove, since start and end points $v_i$, $v_j$ of each edge $e_{i:j}$ strictly satisfy $i < j$ and next edge $e_{j:k}$ will start from $v_j$ and thus get a larger positional encoding.

Formally, for any input token $t_k$, namely edge $e_{i:j}$ covering elements $c_{i+1:j}$, positional encoding $p_{i+1}$ will be used to represent its position and be added to its embedding.


\begin{table*}[t!]
\centering
\begin{tabular}{l|c|l|lllll|l}
      \textbf{System} & \textbf{Input} & \textbf{MT05} & \textbf{MT02} & \textbf{MT03} & \textbf{MT04} & \textbf{MT06} & \textbf{MT08} & \textbf{ALL}\\
  \hline\hline
  \multirow{3}{*}{RNN} & PKU & 31.42 & 34.68 & 33.08 & 35.32 & 31.61 & 23.58 & 31.76 \\
   & CTB & 31.38 & 34.95 & 32.85 & 35.44 & 31.75 & 23.33 & 31.78 \\
   & MSR & 29.92 & 34.49 & 32.06 & 35.10 & 31.23 & 23.12 & 31.35 \\ \hline
  Lattice-RNN  & Lattice & 32.40 & 35.75 & 34.32 & 36.50 & 32.77 & 24.84 & 32.95 \\ \hline
  \multirow{3}{*}{Transformer} & PKU & 41.67 & 43.61 & 41.62 & 43.66 & 40.25 & 31.62 & 40.24 \\
  & CTB & 41.87 & 43.72 & 42.11 & 43.58 & 40.41 & 31.76 & 40.35 \\
   & MSR & 41.17 & 43.11 & 41.38 & 43.60 & 39.67 & 31.02 & 39.87 \\ \hline
   Transformer + LPE & \multirow{3}{*}{Lattice} & 42.37 & 43.71 & 42.67 & 44.43 & 41.14 & 32.09 & $40.93^{\uparrow}$ \\
   Transformer + LSA &  & 42.28 & 43.56 & \textbf{42.73} & 43.81 & 41.01 & 32.39 & $40.77^{\uparrow}$ \\
       Transformer + LPE + LSA &  & \textbf{42.65} & \textbf{44.14} & 42.24 & \textbf{44.81} & \textbf{41.37} & \textbf{32.98} & $\textbf{41.26}^{\uparrow}$ \\
\end{tabular}
\caption{Evaluation of translation performance on NIST Zh-En dataset. RNN and Lattice-RNN results are from \cite{su2017lattice}. We highlight the highest BLEU score in bold for each set. $\uparrow$ indicates statistically significant difference ($p < $0.01) from best baseline.
  }
  \label{LDC}
\end{table*}

\paragraph{Lattice-aware Self-Attention (LSA)} 

We also directly modify self-attention to a lattice-aware way which makes self-attention aware of the relations between any two different edges.  We modified Equations (\ref{1}) and (\ref{3}) in the same way of \citet{shaw2018self} to indicate edge relation:

\begin{eqnarray}
     e_{ij} &=& \frac{(t_iW^Q)(t_jW^K+r^K_{ij})^T}{\sqrt{d}}, \\
     o_i &=& \sum^M_{j=1}\alpha_{ij}(t_jW^V + r^V_{ij}),
\end{eqnarray}%
where $r^K_{ij}$ and $r^V_{ij}$ are relation embeddings which are added to the keys and values to indicate relation between input tokens $t_i$ and $t_j$, namely edges $e_{p:q}$ and $e_{k:l}$ in lattice graph, respectively. 

To facilitate parallel computation, we add an additional embedding (\textbf{self}) for a token when it is conducted dot-product attention with itself, so we train eight (seven in Table \ref{relation}) different relation embeddings $a_{1:8}^V$ and $a_{1:8}^K$ as look-up table for keys and values, respectively. $r^K_{ij}$ and $r^V_{ij}$ can look up for $a_{1:8}^V$ and $a_{1:8}^K$ based on the relation between $t_i$ and $t_j$.
Figure \ref{lpe} shows an example of embeddings in lattice-aware self-attentions based on the timestep of token \emph{fa-zhan-ju} and \emph{fu}.

Since self-attention is computed parallelly, we generate a matrix with all lattice embeddings in it for each sentence which can be easily incorporated into standard self-attention by matrix multiplication. We use different relation embeddings for different Transformer layers but share the same one between different heads in a single layer.

\section{Experiments}

\subsection{Setup}

We conducted experiments on the NIST Chinese-English (Zh-En) and IWSLT 2016 English-German (En-De) datasets. The Zh-En corpus consists of 1.25M sentence pairs and the En-De corpus consists of 191K sentence pairs. For Zh-En task, we chose the NIST 2005 dataset as the validation set and the NIST 2002, 2003, 2004, 2006, and 2008 datasets as test sets. For En-De task, tst2012 was used as validation set and tst2013 and tst2014 were used as test sets. For both tasks, sentence pairs with either side longer than 50 were dropped. We used the case-sensitive 4-gram NIST BLEU score \cite{papineni2002bleu} as the evaluation metric and sign-test \cite{collins2005clause} for statistical significance test.

For Zh-En task, we followed \citet{su2017lattice} to use the toolkit\footnote{https://nlp.stanford.edu/software/segmenter.html\#Download} to train segmenters on PKU, MSR \cite{emerson2005second}, and CTB corpora \cite{xue2005penn}, then we generated word lattices with different segmented training data. Both source and target vocabularies are limited to 30K. For En-De task, we adopted 8K, 16K and 32K BPE merge operations \cite{sennrich2015neural} to get different segmented sentences for building subword lattices. 16K BPE merge operations are employed on the target side.


We set batch size to 1024 tokens and accumulated gradient 16 times before a back-propagation. During training, we set all dropout to 0.3 and chose the Adam optimizer \cite{kingma2014adam} with $\beta_1 = 0.9$, $\beta_2 = 0.98$ and $\epsilon = 10^{-9}$ for parameters tuning. During decoding, we used beam search algorithm and set the beam size to 20. All other configurations were the same with \citet{vaswani2017attention}. We implemented our model based on the OpenNMT \cite{klein2017opennmt} and trained and evaluated all models on a single NVIDIA GeForce GTX 1080 Ti GPU.

\subsection{Overall Performance}


From Table \ref{LDC}, we see that our LPE and LSA models both outperform the Transformer baseline model of 0.58  and 0.42  BLEU respectively. When we combine LPE and LSA together, we get a gain of 0.91 BLEU points. Table \ref{IWSLT} shows that our method also works well on the subword level.

The base Transformer system has about 90M parameters and our LPE and LSA models introduce 0 and 6k parameters over it, respectively, which shows that our lattice approach improves Transformer with little parameter accumulation.

During training, base Transformer performs about 0.714 steps per second while LPE + LSA model can process around 0.328. As lattice-based method usually seriously slows down the training, our lattice design and implementation over the Transformer only shows moderate efficiency reduction.

\begin{table}[t] \small
    \centering
    \begin{tabular}{l|c|l|ll}
    \textbf{System} & \textbf{Input} & \textbf{tst2012} & \textbf{tst2013} & \textbf{tst2014} \\ \hline \hline
    RNN  &16k&26.24&28.22&24.17  \\ \hline
    \multirow{3}{*}{Transformer}&8k& 27.31&29.56&25.57      \\
         & 16k & 27.35 &29.02&25.12 \\
         & 32k &27.15 &28.61&24.88\\ \hline
    + LPE & \multirow{3}{*}{Lattice} & 27.34 &29.48&$25.88^{\uparrow}$ \\
    + LSA &&27.44 &$29.73^{\uparrow}$ &25.65\\
    + LPE + LSA &&\textbf{27.76} & $\textbf{30.28}^{\uparrow}$ &$\textbf{26.22}^{\uparrow}      $
    \end{tabular}
    \caption{Evaluation of translation performance on IWSLT2016 En-De dataset. RNN results are reported from \citet{morishita2018improving}. $\uparrow$ indicates statistically significant difference ($p < $0.01) from best baseline.}
    \label{IWSLT}. 
\end{table}

\subsection{Analysis\footnote{All analysis experiments conducted on NIST dataset.}}

\noindent \textbf{Effect of Lattice-Based Encoders} To show the effectiveness of our method, we placed all edges in the lattice of a single sequence in a relative right order based on their first character, then we applied normal positional encodings (PE) to the lattice inputs on our base Transformer model. As shown in Table \ref{analysis}, our LPE and LSA method outperforms normal positional encodings by 0.39 and 0.23 BLEU respectively which shows that our methods are effective.

\noindent \textbf{Complementary of LPE and LSA} Our LPE method allows edges in all paths in an increasing positional encoding order which seems to focus on long-range order but ignore local disorder. While our LSA method treats all preceding and succeeding edges equally which seems to address local disorder better but ignore long-range order. To show the complementary of these two methods, we also placed all edges of lattice in a single sequence in a relative right order based on their first character and use normal positional encodings and our LSA method; we obtained a BLEU of 40.90 which is 0.13 higher than single LSA model. From this, we can see that long-range position information is indeed beneficial to our LSA model. 

\begin{table}[t] \small
    \centering
    \begin{tabular}{c|c|c} 
    \toprule[1pt]
    \textbf{Systems} & \textbf{ PE}  & \textbf{PE + LSA} \\ \hline
    \textbf{ALL} & 40.54& 40.90  \\
    \bottomrule[1pt]
    \end{tabular}
    \caption{Translation performance (BELU score) with normal positional encodings and normal positional encodings with LSA model on NIST Zh-En dataset.}
    \label{analysis}
\end{table}

\section{Related Work}
Neural network based methods have been applied to several natural language processing tasks \cite{li-etal-2018-seq2seq, zhang2019ime, AAAI1816060,chen-etal-2017-neural, DBLP:journals/corr/abs-1901-05280,he-etal-2018-syntax,zhou-2019-HPSG}, especially to NMT \cite{Bahdanau-EtAl:2015:ICLR2015,wang-etal-2017-sentence,wang-etal-2017-instance,wang-etal-2018-dynamic,8360031,zhang-etal-2018-exploring,Minimum}.

Our work is related to the source side representations for NMT. Generally, the NMT model uses the word as a basic unit for source sentences modeling. In order to obtain better source side representations and avoid OOV problems, recent research has modeled source sentences at character level \cite{DBLP:journals/corr/LingTDB15,costa2016character,yang-etal-2016-character,DBLP:journals/corr/LeeCH16}, subword level \cite{sennrich2015neural,regularization,wu2018find} and mixed character-word level \cite{luong2018open}. All these methods show better translation performance than the word level model.

As models mentioned above only use 1-best segmentation as inputs, lattice which can pack many different segmentations in a compact form has been widely used in statistical machine translation (SMT) \cite{xu2005integrated,dyer2008generalizing} and RNN-based NMT \cite{su2017lattice,sperber2017neural}. To enhance the representaions of the input, lattice has also been applied in many other NLP tasks such as named entity recognition \cite{yue2018chinese}, Chinese word segmentation \cite{yang2018subword} and part-of-speech tagging \cite{jiang2008word,wang2013lattice}.

\section{Conclusions}

In this paper, we have proposed two methods to incorporate lattice representations into Transformer. Experimental results in two datasets on word-level and subword-level respectively validate the effectiveness of the proposed approaches.

Different from \citet{velivckovic2017graph}, our work also provides an attempt to encode a simple labeled graph into Transformer and can be used in any tasks which need Transformer encoder to learn sequence representation. 


\bibliography{acl2019}
\bibliographystyle{acl_natbib}

\end{CJK*}
\end{document}